\documentclass[sigconf ]{acmart}
\usepackage{multirow}
\AtBeginDocument{%
  }

\copyrightyear{2025}
\acmYear{2025}
\setcopyright{cc}
\setcctype{by-nc-nd}
\acmConference[PoIDS '25]{The 1st ACM SIGSPATIAL International Workshop on Polar Data Science}{November 3--6, 2025}{Minneapolis, MN, USA}
\acmBooktitle{The 1st ACM SIGSPATIAL International Workshop on Polar Data Science (PoIDS '25), November 3--6, 2025, Minneapolis, MN, USA}
\acmDOI{10.1145/3764922.3771202}
\acmISBN{979-8-4007-2185-4/2025/11}

\begin{document}

\title{Ice-FMBench: A Foundation Model Benchmark for Sea Ice Type Segmentation}


\author{Samira Alkaee Taleghan}
\affiliation{%
  \institution{University of Colorado Denver}
  \city{Denver}
  \state{Colorado}
  \country{USA}}
\email{samira.alkaeetaleghan@ucdenver.edu}

\author{Morteza Karimzadeh}
\affiliation{%
  \institution{University of Colorado Boulder}
  \city{Boulder}
  \state{Colorado}
  \country{USA}}
\email{karimzadeh@colorado.edu}

\author{Andrew P. Barrett}
\affiliation{%
  \institution{National Snow and Ice Data Center (NSIDC), CIRES, University of Colorado Boulder}
  \city{Boulder}
  \state{CO}
  \country{USA}
}
\email{andrew.barrett@colorado.edu}
\author{Walter N. Meier}

\affiliation{%
  \institution{National Snow and Ice Data Center (NSIDC), CIRES, University of Colorado Boulder}
  \city{Boulder}
  \state{CO}
  \country{USA}
}
\email{walt@colorado.edu}

\author{Farnoush Banaei-Kashani}
\affiliation{%
  \institution{University of Colorado Denver}
  \city{Denver}
  \state{Colorado}
  \country{USA}}
\email{farnoush.banaei-kashani@ucdenver.edu}


\renewcommand{\shortauthors}{Alkaee Taleghan et al.}

\begin{abstract}
Accurate segmentation and mapping of sea ice types is crucial for safe polar navigation, offshore operations, climate monitoring, and ecosystem analysis. While deep learning has demonstrated strong potential for automating sea ice type segmentation, its success often relies on access to extensive expert-labeled datasets, which is both resource-intensive and time-consuming to create. However, foundation models (FMs), recently developed through self-supervised training on large-scale datasets, have demonstrated impressive performance across a range of remote sensing downstream tasks. Nevertheless, their applicability to sea ice type segmentation based on Synthetic Aperture Radar (SAR) imagery remains uncertain due to the unique challenges posed by sea ice—such as intricate geophysical patterns, pronounced seasonal variability, and SAR-specific artifacts like banding, scalloping, and heterogeneous backscatter—as well as the fact that SAR data in polar regions are often acquired using specialized sensor modes that differ markedly from those used to collect FM training data at lower latitudes, limiting their direct transferability to polar environments. To address this gap, we contribute: (1) \textit{Ice-FMBench}, a comprehensive benchmark framework for evaluation of the state-of-the-art remote sensing FMs on the sea ice type segmentation task using Sentinel-1 SAR imagery, where Ice-FMBench is composed of a widely used standardized dataset, diverse evaluation metrics, and a representative set of selected remote sensing FM models most suitable for sea ice type segmentation, with the ability to include new models side-by-side the existing models; (2) an extensive comparative evaluation of the representative FMs using Ice-FMBench, with additional case studies to assess performance of the top-performing model in terms of transferability across temporal and spatial domains and sensitivity to training dataset size; and (3) a multi-teacher knowledge distillation approach to address lack of spatiotemporal transferability of the existing FMs by transferring insight from spatially and temporally specialized expert models into a single, efficient student model.

\end{abstract}


\received{20 February 2007}
\received[revised]{12 March 2009}
\received[accepted]{5 June 2009}

\maketitle

\section{Introduction}
Sea ice is a vital component of the global climate system, and mapping ice types from satellite imagery is crucial for navigation, resource management, and understanding climate dynamics \cite{Mori2019}. Identifying and classifying ice types through segmentation provides crucial information for operational forecasting and climate change monitoring\cite{Vihma2014, Bobylev2020}. Traditionally, specialists relied on manual or rule-based techniques to analyze SAR images, usually integrating auxiliary data like weather records. While valuable, these methods are time-consuming, labor-intensive, and prone to subjectivity \cite{Zakhvatkina, Dedrick}. Growing demand for large-scale, frequent ice monitoring underscores the need for automated segmentation techniques.

For automated monitoring of sea ice, SAR is a frequently chosen sensor due to its ability to image day and night and through all weather conditions, which is vital for consistent observation in polar areas. However, it presents challenges like speckle noise, heterogeneous backscatter, and sensor-specific artifacts that complicate accurate segmentation \cite{Park2020, Dierking2013}. Beyond sea ice segmentation, SAR is extensively utilized in polar regions for applications such as glacier monitoring, permafrost mapping, and iceberg detection, underscoring the broader utility of Ice-FMBench for diverse SAR-based polar remote sensing tasks \cite{Winsvold2018, Mazur2017}. Furthermore, seasonal and regional variations in sea ice introducing additional complexity to model generalization and transferability. 

While traditional machine learning offer some automation, their effectiveness is limited by the scarcity of high-quality labeled data. Deep learning has improved sea ice type segmentation, with CNNs, particularly U-Net\cite{Ronneberger} and its variants, showing strong performance by learning hierarchical features from SAR imagery\cite{Zhang2022, Lima2023}; however, they still require large, high-quality labeled datasets, which are difficult to obtain in polar regions. FMs pre-trained on diverse unlabeled data \cite{Bommasani2021} and fine-tuned for specific tasks have shown strong performance in remote sensing applications\cite{Guo2024, Sun2022, Cha2023, Lu2024}, suggesting potential for polar domains. However, the dynamic, complex nature of sea ice—driven by melt cycles and motion—poses unique challenges to directly applying existing FMs. 

Despite growing interest in remote sensing FMs, no benchmark exists for evaluating their performance in sea ice type segmentation, limiting the development of effective FM-based solutions. While some efforts have introduced benchmarking frameworks for deep learning models in sea ice type \cite{Alkaee2025}, none have focused specifically on modern FMs. Existing benchmarks \cite{Fibaek2024,GEO-Bench,FOMOBench} evaluate FMS on general remote sensing tasks, primarily using optical or multi-modal data in non-polar regions. In contrast, our work introduces Ice-FMBench, a SAR-only, Arctic-focused benchmark tailored specifically for sea ice type segmentation, filling a key gap in FM evaluation for polar applications. 

This work makes the following key contributions: (1) We introduce Ice-FMBench, a comprehensive benchmark framework for evaluating state-of-the-art remote sensing FMs on sea ice type segmentation using Sentinel-1 SAR imagery. Ice-FMBench includes a widely used standardized dataset, diverse evaluation metrics, and a representative selection of FM models, with the flexibility to incorporate new models alongside existing ones for consistent comparison; (2) We conduct an extensive comparative evaluation using Ice-FMBench, including case studies analyzing the spatial and temporal transferability of the top-performing model and its sensitivity to training data size; and (3) We propose a multi-teacher knowledge distillation approach that improves generalization by distilling knowledge from spatially and temporally specialized expert models into a single, efficient student model.

The paper is organized as follows: Section 2 reviews related work; Section 3 describes the benchmark framework; Section 4 presents experiments and results; Section 5 details the knowledge distillation approach; and Section 6 concludes with future directions.
\section{Related Work}
We first review remote sensing FMs, then cover deep learning methods for sea ice type segmentation, and finally discuss relevant benchmarking efforts in both domains.
\vspace{-3mm}
\subsection{Remote Sensing Foundation Models}
FMs learn transferable spatial, spectral, and temporal representations from large datasets\cite{Bommasani2021}, making them valuable for remote sensing where labeled data is scarce \cite{Xiao2024, Alkaee2024}. 
Their development has closely followed advances in pre-training. Early FMs relied on supervised learning using annotated datasets, but their generalization was limited by scarce labels. Self-supervised learning (SSL) addressed this limitation by enabling models to learn from large unlabeled imagery. Among SSL methods, contrastive learning (CL) enables discriminative feature learning across sensors\cite{Fuller2024, Tian2024, Mai2023}, while masked image modeling (MIM) methods like masked autoencoders (MAE) \cite{He2022} enable contextual understanding and robust feature learning \cite{MMEarth, Jakubik2023, Cong2022, Tang2023, Wang2024, Li2025, Wang2022, Xiong2024}. Recent hybrid approaches such as Contrastive Masked Image Distillation (CMID) \cite{Muhtar2023} combine CL and MIM for improved representation learning.

Beyond pre-training strategies, multi-modal data improves FM robustness and generalization. Modalities like Multispectral Imaging (MSI), Hyperspectral Imaging (HSI), SAR, and LiDAR provide complementary geospatial information. Including these during pre-training enhances robustness to seasonal variation and sensor noise \cite{Wang2022b, Xiong2024}. Datasets such as BigEarthNet\cite{BigEarthNet}, SSL4EO-S12 \cite{SSL4EO} incorporate Sentinel-1 SAR to capture radar-specific features. However, many FMs still rely mainly on RGB data \cite{Muhtar2023, Wang2022, Wang2023}, such as MillionAID \cite{MillionAID}, limiting their effectiveness for SAR tasks. Recent studies have tried to address this gap. Li et al. \cite{Li2025} focused on SAR target recognition, while Guo et al. \cite{Guo2024} aligned RGB, MSI, and SAR data via CL. However, neither used SAR frequencies or polarizations specific to sea ice mapping.
Semantic segmentation for FMs leverages Transformer-based models like Vision Transformers (ViTs) \cite{vit} and Swin to capture long-range dependencies via self-attention \cite{ Manas2021, Prexl2023, Wang2024, Satlas}. 

While FMs have advanced remote sensing, their application to sea ice type segmentation is still limited. Most are pre-trained on diverse earth observation data but often exclude polar-relevant modalities and dynamics. In particular, models incorporating Sentinel-1 data have largely relied on lower-latitude imagery acquired in the Interferometric Wide Swath (IW) mode, while over polar oceans, Sentinel-1 operates in the Extra Wide (EW) Swath mode, which differs in spatial resolution, polarization, and noise characteristics which is crucial for detecting ice features like texture and surface roughness. Additionally, the dynamic nature of sea ice (drift, melt, refreeze) limits temporally consistent views, posing a challenge for self-supervised models that rely on multiple observations of the same target. As a result, FM generalization in polar environments remains largely untested and needs systematic evaluation.
\vspace{-3mm}
\subsection{Sea Ice Type Segmentation Methods}
Sea ice type segmentation classifies each pixel in satellite imagery into distinct ice types \cite{Minaee2021}. U-Net and its variants are widely used for their strong spatial localization, originally developed for biomedical imaging \cite{Ronneberger} and later adapted for sea ice tasks. It has proven effective in distinguishing ice from open water \cite{Ren2020, Wang2021} and in segmenting multiple ice types under varying conditions \cite{Huang2021}. Multitask U-Nets with downscaling and spatio-temporal encoding have segmented up to six ice types \cite{Chen2024, Chen2023, Chen2024x}. Architectural improvements include dual-attention mechanisms \cite{Ren2021}, multi-task learning \cite{Cantu}, and pre-trained backbones like ResNet50 and VGG-16 to enhance feature extraction \cite{Cantu, Zhao2022}, addressing SAR-specific challenges with improved robustness.
In addition to U-Net-based models, DeepLab-based models have gained attention for their ability to capture multi-scale spatial patterns using atrous convolutions \cite{Chen, Chen2017}. DeepLabv3 combined with ResNet backbones and Atrous Spatial Pyramid Pooling (ASPP) has shown improved segmentation accuracy over U-Net baselines \cite{Lima2023, Jalayer2025}. Enhancements include coordinate attention \cite{Sun2023} and attention-based decoders for better delineation of complex ice structures \cite{Zhang2022}. 
While deep learning has significantly advanced sea ice segmentation, challenges remain. Many methods rely on large labeled datasets, which are scarce in polar regions, and often trade off fine-grained accuracy for robustness. 
\subsection{Model Benchmarking}
Recent benchmarks aim to standardize FM evaluation in remote sensing. PhilEO Bench targets Sentinel-2 for land cover, buildings, and roads \cite{Fibaek2024}; FoMo-Bench focuses on forest monitoring \cite{FOMOBench}; GEO-Bench covers 12 tasks with diverse modalities, mainly from temperate regions \cite{GEO-Bench}; and Copernicus Bench spans 15 tasks across all major Sentinel missions \cite{CopernicusBench}. However, these benchmarks focus on non-polar, optical, terrestrial applications and lack polar-specific tasks. To address the domain-specific requirements of sea ice applications, specialized frameworks like IceBench \cite{Alkaee2025} provide a standardized benchmark for deep learning-based sea ice type segmentation and classification, but do not evaluate modern FMs or their transferability in polar settings. This gap motivates our work, while general FM benchmarks lack polar-specific evaluation and sea ice benchmark exclude FMs, there is no systematic assessment of how state-of-the-art FMs perform on sea ice type segmentation.

\section{Ice-FMBench}
This section introduces Ice-FMBench, the benchmark framework for evaluating FMs in sea ice type segmentation. Ice-FMBench consists of three core components: a dataset, selected representative remote sensing FMs, and standardized evaluation metrics. The benchmark framework is available at: \url{https://github.com/UCD-BDLab/Ice-FMBench}.
\vspace{-3mm}
\subsection{Dataset}
A reliable FM benchmark for sea ice segmentation requires a comprehensive dataset. We use the AI4Arctic Sea Ice Challenge Dataset \cite{ChallengeDataset} as the foundation of Ice-FMBench due to its Sentinel-1 SAR imagery, wide coverage, and established use in community challenges. Originally developed for the AutoICE competition \cite{autoice}, it provides C-band SAR scenes with manual ice charts from the Danish Meteorological Institute (DMI) and the Canadian Ice Service (CIS). Two versions are provided: a raw format for custom processing and a Ready-to-Train (RTT) format that simplifies data preparation. We adopt the RTT version, allocating 512 files for training and reserving a distinct set of 20 files for testing.

The selection of the AI4Arctic dataset for Ice-FMBench is motivated by several key strengths. It covers January 2018–December 2021 across 16 regions, it captures seasonal and regional sea ice variability, enabling FM fine-tuning and testing generalization over time and space. Its prior use in the AutoICE competition further establishes it as a recognized benchmark within the research community, enabling meaningful comparisons with existing methods and encouraging broader participation. The dataset is publicly available, which supports reproducibility and promotes continued research. It is important to note that the dataset is limited to Greenland waters and does not represent the full Arctic region; comparable datasets for the Southern Ocean remain unavailable.

\begin{table*}[t]
\centering
\caption{Summary of Remote Sensing FMs}
\vspace{-3mm}
\label{tab:model_summary}
\footnotesize
\setlength{\tabcolsep}{4pt}
\renewcommand{\arraystretch}{1.1}
\begin{tabular}{@{}l 
                  p{1.2cm}
                  p{1.5cm} 
                  c 
                  p{1.3cm} 
                  l 
                  p{2.5cm}
                  c
                  p{1.2cm}
                  c@{}}
\toprule
\textbf{Model} & 
\textbf{\# Params. (M)} & 
\textbf{Pre-training Dataset} & 
\textbf{Res. (m)} & 
\textbf{Coverage} & 
\textbf{Visual Encoder} & 
\textbf{Pretrain Method} &
\textbf{Input Modalities} &
\textbf{Code Available} &
\textbf{Year} \\
\midrule
Prithvi-100m \cite{Jakubik2023}  & 100    & HLS & 30 & U.S & ViT & MAE  & MSI & \checkmark & 2023 \\
Prithvi-300m \cite{Szwarcman2024}  & 300   & HLS & 30 & Global & ViT & MAE  & MSI & \checkmark & 2024 \\
Prithvi-600m \cite{Szwarcman2024} & 600   & HLS & 30 & Global & ViT & MAE & MSI & \checkmark & 2024 \\
\addlinespace
CROMA \cite{Fuller2024}        & 159   & SSL4EO-S12 & 10/30/60 & Global & Multi-Modal ViT & CL + Multimodal MAE & SAR, MSI & \checkmark & 2023 \\
DINO-MM \cite{Wang2022b}       & 21.7  & BigEarthNet & 10 & Europe & ViT-DINO & DINO distillation & SAR, MSI & \checkmark & 2022 \\
DOFA \cite{Xiong2024}         & 337   & Sentinel-1/2, Gaofen, NAIP,  & 1–30 & Global & ViT & MIM + Distillation & SAR, MSI,RGB,etc & \checkmark & 2024 \\
CMID \cite{Muhtar2023}         & 86.7  & MillionAID & 10 & Global & Swin, ResNet50 & Contrastive Mask Image Distillation  & RGB & \checkmark & 2023 \\
SARATR-X \cite{Li2025}     & 66 & SARDet-180K &  0.3-25 & Global & HiViT & Two-step MAE with MGF & SAR  & \checkmark & 2025 \\
FG-MAE \cite{Wang2024}         & 304   & SSL4EO-S12 & 10 & Global & ViT & Feature-Guided MAE & SAR, MSI & \checkmark & 2023 \\
RVSA \cite{Wang2022}         & 100  & MillionAID & 0.5-11.4 & Global & ViT, ViTAE + RVSA & Unsupervised MAE & RGB & \checkmark & 2022 \\
SSL4EO \cite{SSL4EO}      & 85.6  & SSL4EO-S12 & 10/30/60 & Global & ViT & MAE & SAR, MSI & \checkmark & 2023 \\
\bottomrule
\end{tabular}
\vspace{-2mm}
\end{table*}
\vspace{-4mm}
\subsection{Methods}
Selecting appropriate FMs is key to making Ice-FMBench a reliable benchmark for sea ice type segmentation. Models were chosen based on: (1) SAR relevance, including those trained or evaluated on Sentinel-1 SAR—either SAR-specific or multi-modal—to handle speckle noise and backscatter variability; (2) benchmark performance, favoring models that performed well on tasks like ISPRS Potsdam \cite{Rottensteiner2012}, particularly those using multi-scale, attention, or transformer-based architectures \cite{Lu2024}; and (3) self-supervised learning pretraining, including CL and MIM, which are effective in label-scarce Arctic settings \cite{Lu2024}. Based on these criteria, we selected eleven diverse FMs with available implementations and pre-trained weights. A full summary is shown in Table \ref{tab:model_summary}.

Among FMs, RVSA \cite{Wang2022} and CMID \cite{Muhtar2023} have achieved state-of-the-art results on benchmarks like ISPRS Potsdam. RVSA is noted for its exceptional overall accuracy, while CMID excels in IoU scores, indicating robust delineation of class boundaries \cite{Lu2024}. CMID employs a unified SSL framework that integrates CL and MIM to capture both global semantic and local spatial representations, using ResNet-50 or Swin Transformer backbones \cite{Muhtar2023}. Similarly, RVSA enhances ViTs \cite{vit} through a Rotated Varied-Size Attention (RVSA) mechanism with learnable rotation, alongside MAE pre-training, yielding competitive results on datasets like iSAID and Potsdam \cite{Wang2022, Zamir2019}. Both, however, were pre-trained solely on optical data, limiting SAR adaptation.
The Prithvi family of models \cite{Jakubik2023, Szwarcman2024} utilizing a ViT backbone with MAE pre-training. Prithvi-EO-1.0(100M parameters) was pre-trained on Harmonized Landsat-Sentinel 2 (HLS) data from the U.S., while the larger Prithvi-EO-2.0 variants (300M and 600M parameters) used global HLS data and incorporated temporal embeddings and metadata for enhanced spatiotemporal learning. Prithvi-EO-2.0 demonstrates strong segmentation performance, ranks highly on GEO-Bench \cite{GEO-Bench}, and has been applied to diverse tasks such as flood mapping and wildfire detection.

Models integrating multi-modal or SAR data integration show strong potential for sea ice tasks. DINO-MM \cite{Wang2022b} extends DINO \cite{Caron2021} to jointly learn SAR and optical features using self-supervised transformers, with RandomSensorDrop to improve cross-modal robustness. It is trained on BigEarthNet \cite{BigEarthNet}, which includes Sentinel-1 SAR. CROMA \cite{Fuller2024} uses CL with ViT backbones to fuse Sentinel-1 SAR (VV/VH) and Sentinel-2 MSI, employing advanced positional encoding strategies (X-ALiBi and 2D-ALiBi \cite{Press2021}) for large-image processing, and achieves top results on DFC2020 \cite{Yokoya2020}. FG-MAE \cite{Wang2024}, a ViT-based MAE adapted for SAR, leverages HOG \cite{Dalal2005} to enhance spatial features and reduce speckle, demonstrating strong segmentation transferability. Similarly, the SSL4EO \cite{SSL4EO} applied SSL frameworks such as MoCo-v2, DINO, MAE, and data2vec through large-scale pretraining on multi-modal Sentinel-1/2 data.

DOFA \cite{Xiong2024} applies MIM on 4.6M Sentinel-1 SAR and other modalities with a dynamic hypernetwork for sensor adaptation, achieving state-of-the-art SegMunich \cite{Hong2024} performance. SARATR-X \cite{Li2025}, designed for SAR ATR, uses a Hierarchical ViT(HiViT) backbone with two-step pre-training process involves an ImageNet-based MIM approach followed by SAR-specific MIM using multi-scale gradient features (MGFs) to reduce speckle and refine targets.
This diverse FM ensemble allows Ice-FMBench to assess how architectures and SAR integration strategies handle multi-scale context, speckle noise, and effectiveness in sea ice type segmentation.

\subsection{Metrics}
Ice-FMBench evaluates FMs for sea ice segmentation using two metrics: (1) accuracy, measuring agreement with ground truth labels, and (2) efficiency, capturing computational cost during training and inference.

\vspace{-2mm}
\subsubsection{Accuracy Metrics}
We evaluate multiclass, per-pixel metrics and report frequency-weighted (support-weighted) averages across ice-type classes. 
Let $\mathcal{C}=\{1,\dots,K\}$ be the set of classes. 
For each class $c\in\mathcal{C}$, let $\mathrm{TP}_c,\mathrm{FP}_c,\mathrm{FN}_c$ be pixel counts, 
$n_c=\mathrm{TP}_c+\mathrm{FN}_c$ the number of ground-truth pixels, and 
$w_c=\frac{n_c}{\sum_{j\in\mathcal{C}} n_j}$ the frequency weight. 
Per-class metrics are:
\[
\begin{aligned}
\mathrm{Prec}_c &= \tfrac{\mathrm{TP}_c}{\mathrm{TP}_c+\mathrm{FP}_c}, \quad &
\mathrm{Rec}_c  &= \tfrac{\mathrm{TP}_c}{\mathrm{TP}_c+\mathrm{FN}_c},\\
\mathrm{F1}_c   &= \tfrac{2\,\mathrm{TP}_c}{2\,\mathrm{TP}_c+\mathrm{FP}_c+\mathrm{FN}_c}, \quad &
\mathrm{IoU}_c  &= \tfrac{\mathrm{TP}_c}{\mathrm{TP}_c+\mathrm{FP}_c+\mathrm{FN}_c}.
\end{aligned}
\]

We report:
\[
\mathrm{Accuracy}=\frac{\sum_{c\in\mathcal{C}}\mathrm{TP}_c}{\sum_{c\in\mathcal{C}} n_c},
\]
\[
\mathrm{Precision}=\sum_{c\in\mathcal{C}} w_c\,\mathrm{Prec}_c,\qquad
\mathrm{Recall}=\sum_{c\in\mathcal{C}} w_c\,\mathrm{Rec}_c,
\]
\[
\mathrm{F1}=\sum_{c\in\mathcal{C}} w_c\,\mathrm{F1}_c=\sum_{c\in\mathcal{C}} w_c\,\frac{2\,\mathrm{Prec}_c\,\mathrm{Rec}_c}{\mathrm{Prec}_c+\mathrm{Rec}_c},\qquad
\mathrm{IoU}=\sum_{c\in\mathcal{C}} w_c\,\mathrm{IoU}_c.
\]

\subsubsection{Efficiency Metrics}
To assess the operational cost of each FM, we record the following average efficiency metrics during both training and inference phases.
 \textit{Memory Consumption:} We monitor GPU and RAM usage (in GB) to assess hardware memory needs for deployment across processing units.
 \textit{Hardware Utilization:} We track GPU and CPU usage percentage to assess how effectively each model leverages computational resources during training and testing.
 \textit{Processing Time:} We measure training time through average time per epoch (minutes per epoch) and evaluate inference time by recording total processing time for the test dataset (minutes).

\section{Comparative Experimental Analysis}
This section presents the methodology and comparative evaluation of FMs for sea ice type segmentation using Ice-FMBench.
\vspace{-3mm}
\subsection{Experimental Methodology}
For this study, we utilized the RTT version of the AI4Arctic Sea Ice Challenge Dataset \cite{ChallengeDataset}. Our experiments use Sentinel-1 C-band SAR EW-mode Ground Range Detected (GRD) imagery with dual-polarization (HH, HV) at 80 m resolution, collected from 2018–2021. The dataset includes 513 training scenes and 20 test scenes. The ground truth labels for sea ice type segmentation are derived from manually produced ice charts by the DMI and CIS. In the dataset, these are provided as pixel-level Stage of Development (SOD) labels. Sea ice type is categorized into six distinct classes based on SOD: 0 (Open Water), 1 (New Ice), 2 (Young Ice), 3 (Thin First-Year Ice), 4 (Thick First-Year Ice), and 5 (Old Ice). These pixel-level labels originate from manually drawn polygons in the ice charts, which delineate ice conditions. Each polygon is characterized by attributes including SOD and ice concentration, with the dominant ice type assigned to the entire polygon if it covers at least 65\% of its area, forming the basis for the provided segmentation masks.

For training and evaluation, Sentinel-1 SAR images are split into $224 \times 224$ patches; random cropping, flips, and rotations are applied for spatial variability and robustness. To ensure compatibility across models with different input requirements, we adapt the input channels accordingly. Prithvi, RVSA, and CMID (three-channel) use HH, HV, and the ratio of HH to HV (HH/HV), while SAR-specific models (FG-MAE, SARATR-X, DOFA, DINO-MM, CROMA) use HH and HV. All channels are normalized using precomputed statistics. To enhance generalization, we apply data augmentations such as random flips and rotations, simulating natural SAR variability and boosting robustness to diverse sea ice conditions. The backbone architectures used in our experiments include ViT-MAE (Prithvi-100/300/600M), ViT-L (CROMA, RVSA), ViT-S/8 (DINO-MM), ViT-L SAR-pretrained (DOFA), Swin-L (CMID), HiViT-B (SARATR-X), ViT-L/16 SAR-pretrained (FG-MAE), and ViT-B/16 MAE SAR-pretrained (SSL4EO).
As baselines from IceBench \cite{Alkaee2025}, we use U-Net \cite{Huang2021} ([32, 32, 64, 64] channels) and DeepLabV3 \cite{PiresdeLima2023} with a ResNet-18 backbone pretrained on ImageNet \cite{Deng2009}, serving as references for comparing FM segmentation performance.

To assess FM adaptability and transferability to SAR sea ice segmentation, we implement five fine-tuning strategies varying in parameter updates, computational cost, and architecture changes. These strategies span the major categories of parameter-efficient fine-tuning (PEFT) \cite{Houlsby2019} methods—additive (prompt-base), partial, reparameterized tuning—following the taxonomy proposed by Xu et al. \cite{Xu2023} along with two baselines: frozen encoder with trainable decoder and full fine-tuning. These strategies span minimal tuning to full retraining, enabling exploration of performance–efficiency trade-offs in the SAR domain. Based on this categorization, we define five strategies:

  \textit{Strategy 1: VPT (Visual Prompt Tuning )\cite{Jia2022} (Additive).} Inserts learnable prompt tokens into the encoder input for efficient fine-tuning. Using the VPT-Deep variant, prompts are added at every transformer layer, randomly initialized, and trained while keeping encoder weights frozen. The prompt length is set to 10.

  \textit{Strategy 2: BitFit (Bias-only Fine-Tuning)\cite{Zaken2021}(Partial).} Freezes all encoder parameters except bias terms, updating only these and the segmentation decoder. This highly parameter-efficient method supports rapid, low-resource domain adaptation.
  
   \textit{Strategy 3: LoRA (Low-Rank Adaptation)\cite{Hu2021} (Reparameterized).} In this strategy, we inject low-rank matrices into the query and value projections of a frozen backbone’s attention modules for efficient adaptation. Only LoRA parameters (rank 4, alpha 16, dropout 0.2) and the decoder are trained, keeping the backbone frozen.
  
   \textit{Strategy 4: Frozen Encoder with Trainable Decoder.} With this strategy, the encoder remains frozen while only the segmentation decoder is fine-tuned. This is the most computationally efficient setup but may miss domain-specific features.

  \textit{Strategy 5: Full Fine-Tuning.} With this strategy, both encoder and decoder are trainable, enabling full adaptation to SAR and sea ice features. This often yields top performance but with the highest computational cost, serving as an upper-bound baseline.
  
All models use cross-entropy loss with an ignore index of 255 for no-data regions. Training employs AdamW ($1 \times 10^{-4}$ LR) with StepLR (0.9 decay every 10 epochs) and early stopping (patience 20) based on validation loss. Batch size is 32, with validation on 18 scenes and testing on 20 fixed scenes. Experiments were performed using an NVIDIA RTX A6000 GPU.
\vspace{-3mm}
\subsection{Experimental Results}
We present our findings in four parts: (1) performance comparison, (2) efficiency analysis, (3) transferability evaluation, and (4) sensitivity to training data size. 
\begin{figure}[t]
    \centering
    \includegraphics[width=\linewidth]{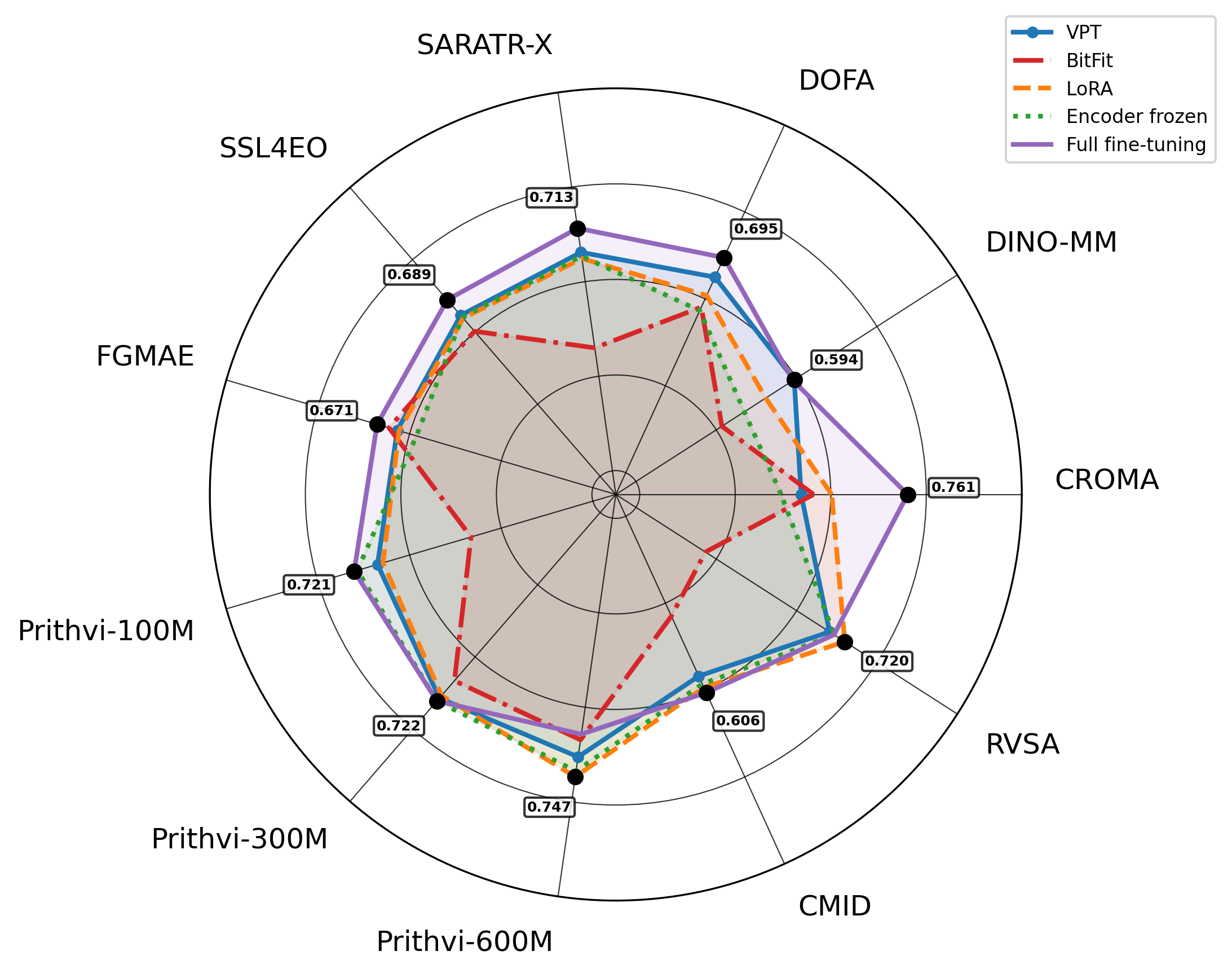}
    \caption{Radar plot of F1-scores across different fine-tuning strategies for remote sensing FMs}
    \Description{Radar chart comparing F1-scores of multiple remote sensing foundation models under five fine-tuning strategies: VPT, BitFit, LoRA, Encoder Frozen, and Full Fine-Tuning. Prithvi-600M and Prithvi-300M achieve the highest overall performance across most strategies. Encoder Frozen and Full Fine-Tuning show consistently higher scores than VPT and BitFit. Models like SARATR-X, DOFA, and CROMA perform moderately, while CMID and RVSA have lower scores across strategies.}
    \vspace{-3mm}
    \label{fig:radar_chart_f1_scores}
\end{figure}
\vspace{-2mm}
\begin{table}[t]
    \centering
    \caption{Accuracy Results of Remote Sensing FMs for Sea Ice Type Segmentation}
    \vspace{-3mm}
    \label{tab:foundation_models_results}
    \setlength{\tabcolsep}{3.5pt}
    \renewcommand{\arraystretch}{0.85}
    \footnotesize
    \begin{tabular}{l c c c c c c}
        \toprule
        \textbf{Model} & \textbf{Channels} & \textbf{F1} & \textbf{Acc.} & \textbf{Prec.} & \textbf{Rec.} & \textbf{IoU} \\
        \midrule
        \multicolumn{7}{@{}l@{}}{\scriptsize\textit{Baseline Models}} \\
        U-Net & 2 & 0.766 & 0.743 & 0.899 & 0.743 & 0.696 \\
        DeepLabV3 & 2 & 0.758 & 0.736 & 0.928 & 0.736 & 0.688 \\
        U-Net & 3 & 0.733 & 0.706 & 0.865 & 0.706 & 0.654 \\
        DeepLabV3 & 3 & 0.714 & 0.690 & 0.840 & 0.690 & 0.642 \\
        \midrule
        \multicolumn{7}{@{}l@{}}{\scriptsize\textit{Strategy 1: VPT}} \\
        CROMA & 2 & 0.538 & 0.594 & 0.858 & 0.594 & 0.488 \\
        DINO-MM & 2 & 0.594 & 0.598 & 0.870 & 0.598 & 0.522 \\
        DOFA & 2 & 0.650 & 0.617 & 0.871 & 0.617 & 0.560 \\
        SARATR-X & 2 & 0.662 & 0.642 & 0.892 & 0.642 & 0.577 \\
        SSL4EO & 2 & 0.646 & 0.646 & 0.852 & 0.646 & 0.575 \\
        FG-MAE & 2 & 0.627 & 0.613 & 0.862 & 0.613 & 0.544 \\
        Prithvi-100M & 3 & 0.670 & 0.645 & 0.862 & 0.645 & 0.577 \\
        \textbf{Prithvi-300M} & 3 & \textbf{0.714} & \textbf{0.672} & \textbf{0.914} & \textbf{0.672} & \textbf{0.634} \\
        Prithvi-600M & 3 & 0.705 & 0.656 & 0.908 & 0.656 & 0.626 \\
        CMID & 3 & 0.568 & 0.558 & 0.755 & 0.558 & 0.465 \\
        RVSA & 3 & 0.682 & 0.689 & 0.876 & 0.688 & 0.610 \\
        
        \midrule
        \multicolumn{7}{@{}l@{}}{\scriptsize\textit{Strategy 2: BitFit}} \\
        CROMA & 2 & 0.564 & 0.547 & 0.804 & 0.547 & 0.470 \\
        DINO-MM & 2 & 0.414 & 0.428 & 0.692 & 0.428 & 0.336 \\
        DOFA & 2 & 0.581 & 0.508 & 0.797 & 0.508 & 0.463 \\
        SARATR-X & 2 & 0.460 & 0.520 & 0.713 & 0.520 & 0.398 \\
        SSL4EO & 2 & 0.602 & 0.620 & 0.868 & 0.620 & 0.536 \\
        FG-MAE & 2 & 0.645 & 0.646 & 0.875 & 0.646 & 0.572 \\
        Prithvi-100M & 3 & 0.466 & 0.559 & 0.858 & 0.559 & 0.420 \\
        Prithvi-300M & 3 & 0.666 & 0.628 & 0.863 & 0.726 & 0.567 \\
        \textbf{Prithvi-600M} & 3 & \textbf{0.669} & \textbf{0.674} & \textbf{0.853} & \textbf{0.791} & \textbf{0.593} \\
        CMID & 3 & 0.430 & 0.368 & 0.699 & 0.368 & 0.305 \\
        RVSA & 3 & 0.373 & 0.286 & 0.766 & 0.286 & 0.258 \\

        \midrule
        \multicolumn{7}{@{}l@{}}{\scriptsize\textit{Strategy 3: LoRA }} \\
        CROMA & 2 & 0.602 & 0.635 & 0.846 & 0.635 & 0.534 \\
        DINO-MM & 2 & 0.523 & 0.584 & 0.789 & 0.584 & 0.471 \\
        DOFA & 2 & 0.608 & 0.570 & 0.845 & 0.570 & 0.502 \\
        SARATR-X & 2 & 0.649 & 0.630 & 0.889 & 0.630 & 0.567 \\
        SSL4EO & 2 & 0.638 & 0.638 & 0.868 & 0.638 & 0.563 \\
        FG-MAE & 2 & 0.623 & 0.610 & 0.870 & 0.610 & 0.544 \\
        Prithvi-100M & 3 & 0.658 & 0.652 & 0.867 & 0.652 & 0.569 \\
        Prithvi-300M & 3 & 0.707 & 0.686 & 0.905 & 0.686 & 0.626 \\
        \textbf{Prithvi-600M} & 3 & \textbf{0.747} & \textbf{0.728} & \textbf{0.933} & \textbf{0.728} & \textbf{0.681} \\
        CMID & 3 & 0.594 & 0.553 & 0.790 & 0.553 & 0.482 \\
        RVSA & 3 & 0.720 & 0.713 & 0.905 & 0.713 & 0.651 \\
        \midrule
        
        \multicolumn{7}{@{}l@{}}{\scriptsize\textit{Strategy 4: Encoder frozen with Trainable Decoder}} \\
        CROMA & 2 & 0.496 & 0.575 & 0.864 & 0.575 & 0.446 \\
        DINO-MM & 2 & 0.470 & 0.561 & 0.862 & 0.561 & 0.424 \\
        DOFA & 2 & 0.573 & 0.590 & 0.838 & 0.590 & 0.497 \\
        SARATR-X & 2 & 0.654 & 0.613 & 0.887 & 0.613 & 0.550 \\
        SSL4EO & 2 & 0.641 & 0.639 & 0.862 & 0.639 & 0.562 \\
        FG-MAE & 2 & 0.585 & 0.593 & 0.835 & 0.593 & 0.504 \\
        Prithvi-100M & 3 & 0.714 & 0.698 & 0.921 & 0.698 & 0.644 \\
        Prithvi-300M & 3 & 0.722 & 0.699 & 0.920 & 0.699 & 0.645 \\
        \textbf{Prithvi-600M} & 3 & \textbf{0.735} & \textbf{0.722} & \textbf{0.929} & \textbf{0.722} & \textbf{0.671} \\
        CMID & 3 & 0.586 & 0.560 & 0.769 & 0.560 & 0.482 \\
        RVSA & 3 & 0.694 & 0.687 & 0.888 & 0.687 & 0.611 \\

        \midrule
        \multicolumn{7}{@{}l@{}}{\scriptsize\textit{Strategy 5: Full fine-tuning}} \\
        \textbf{CROMA} & 2 & \textbf{0.761} & \textbf{0.738} & \textbf{0.940} & \textbf{0.738} & \textbf{0.694} \\
        DINO-MM & 2 & 0.591 & 0.586 & 0.868 & 0.586 & 0.517 \\
        DOFA & 2 & 0.695 & 0.683 & 0.912 & 0.683 & 0.622 \\
        SARATR-X & 2 & 0.713 & 0.687 & 0.914 & 0.687 & 0.632 \\
        SSL4EO & 2 & 0.689 & 0.661 & 0.908 & 0.661 & 0.610 \\
        FG-MAE & 2 & 0.671 & 0.620 & 0.899 & 0.620 & 0.580 \\
        Prithvi-100M & 3 & 0.721 & 0.702 & 0.933 & 0.702 & 0.649 \\
        Prithvi-300M & 3 & 0.722 & 0.711 & 0.922 & 0.711 & 0.650 \\
        Prithvi-600M & 3 & 0.657 & 0.663 & 0.896 & 0.663 & 0.579 \\
        CMID & 3 & 0.606 & 0.589 & 0.827 & 0.589 & 0.506 \\
        RVSA & 3 & 0.693 & 0.697 & 0.903 & 0.697 & 0.624 \\
     
        \bottomrule
    \end{tabular}
    \vspace{-5mm}
\end{table}

\subsubsection{Performance Across Fine-Tuning Strategies}
The performance of the selected remote sensing FMs was evaluated against SOTA deep learning models for sea ice type segmentation, namely, U-Net and DeepLabV3, using five distinct fine-tuning strategies. The F1-scores for fine-tuning strategies across different models on the test set are visualized in Figure \ref{fig:radar_chart_f1_scores}, and detailed accuracy metrics on the test set are presented in Table \ref{tab:foundation_models_results}.
\textit{VPT}: Prithvi-300M (3 channels) achieves the highest F1-score of 0.714 with this strategy, followed by Prithvi-600M (0.705). Prithvi-100M also shows a competitive F1 of 0.670. The radar plot (purple line) shows VPT offering a moderate performance across models, often better than BitFit and frozen encoder approaches for several models but not typically reaching the levels of full fine-tuning or LoRA for the top performers. 

\textit{BitFit}: generally results in lower performance compared to other strategies, as depicted by the red line in the radar plot which is often closer to the center. Prithvi-600M (3 channels) was the top performer under BitFit with an F1-score of 0.669, followed by Prithvi-300M (0.666) and FG-MAE (0.645). Models like DINO-MM (0.414) and RVSA (0.373) showed particularly low F1-scores, suggesting that fine-tuning only bias terms is insufficient for adapting these models effectively to the sea ice segmentation task. 

\textit{LoRA}: shows an improvement over the frozen encoder strategy for most models. Again, Prithvi-600M (3 channels) led with an F1-score of 0.747, closely follow by RVSA (0.720) and Prithvi-300M (0.707). The radar plot (dashed orange line) indicates that LoRA generally provides a balanced performance, often outperforming the frozen encoder approach. For example, CROMA's F1-score improves from 0.496 (Strategy 4) to 0.602 with LoRA.

\textit{Encoder frozen with trainable decoder}: the Prithvi models demonstrate superior performance, particularly Prithvi-600M (3 channels), which achieves the highest F1-score of 0.735, follow by Prithvi-300M (0.722) and Prithvi-100M (0.714). Models like CROMA and DINO-MM show lower F1-scores with this strategy, indicating that their pre-trained encoder features were less directly transferable to the sea ice segmentation task without further fine-tuning. As seen in Figure \ref{fig:radar_chart_f1_scores}, the "Encoder frozen" (dotted green line) sits in the mid-to-lower range for most models except the Prithvi series.

\textit{Full fine-tuning}: This strategy yields the highest F1-scores for several models. Notably, CROMA (2 channels) achieved the overall best F1-score of 0.761 across all models and strategies in this category, surpassing even the U-Net baseline. SARATR-X also performs well with an F1-score of 0.713. However, interestingly full fine-tuning is not universally the best for all models; for instance, Prithvi-600M's F1-score drop to 0.657 compared to its performance with LoRA (0.747) or frozen encoder (0.735). This suggests that for some larger models, full fine-tuning might lead to overfitting or disruption of useful pre-trained features when the downstream dataset is limited. 


The Prithvi model family demonstrated superior performance across multiple strategies, with larger variants (300M and 600M parameters) generally outperforming the 100M version. CROMA showed exceptional performance under full fine-tuning, achieving results comparable to traditional baselines, while models like DINO-MM and CMID consistently underperformed across all strategies, indicating potential architectural limitations for this specific domain. CROMA outperforms other SAR models by integrating contrastive and reconstruction objectives, leveraging cross-modal learning with optical data, and employing specialized architectures like 2D-ALiBi for spatial and geometric understanding. The results suggest that model architecture and scale matter more than domain-specific pretraining when computational resources are limited. SAR pretraining only shows its advantage when you can afford to fine-tune the entire model.
Model selection proves critical, with Prithvi-600M using LoRA adaptation delivering the best overall results (F1 = 0.747), while CROMA achieved competitive performance (F1 = 0.761) under full fine-tuning for 2-channel configuration. The substantial performance drops observed with BitFit underscore that bias-only fine-tuning provides insufficient flexibility for adapting to sea ice segmentation complexities. 
FM architectures demonstrated markedly different transferability characteristics. Models like Prithvi and RVSA consistently showed strong adaptation capabilities across multiple strategies, while others such as DINO-MM and CMID struggled with effective domain transfer, suggesting that certain architectural designs may be more conducive to remote sensing applications.
These results indicate that while FMs show promise for sea ice segmentation, careful consideration of fine-tuning strategy and model architecture. The superior performance of LoRA adaptation suggests may be more effective than either freezing encoders or full fine-tuning.

\vspace{-2mm}
\begin{table}[!t]
\renewcommand{\arraystretch}{1.0}
\setlength{\tabcolsep}{1.5pt}
\centering
\caption{Comparison of Efficiency Metrics Across Fine-Tuning Strategies}
\vspace{-3mm}
\label{tab:efficiency_metrics}
\resizebox{\columnwidth}{!}{%
\fontsize{6pt}{7.5pt}\selectfont
\begin{tabular}{@{}l l c c c c c c c c c c@{}}
\toprule
\multirow{3}{*}{\textbf{Model}} &
\multirow{3}{*}{\textbf{Strat.}} &
\multicolumn{5}{c}{\textbf{Training Phase}} &
\multicolumn{5}{c}{\textbf{Inference Phase}} \\
\cmidrule(lr){3-7} \cmidrule(lr){8-12}
 &  & \textbf{GPU Mem} & \textbf{RAM} & \textbf{GPU Util} & \textbf{CPU Util} & \textbf{T-Time}
 & \textbf{GPU Mem} & \textbf{RAM} & \textbf{GPU Util} & \textbf{CPU Util} & \textbf{I-Time} \\
 &  & \textbf{(GB)} & \textbf{(GB)} & \textbf{(\%)} & \textbf{(\%)} & \textbf{(min/ep)}
 & \textbf{(GB)} & \textbf{(GB)} & \textbf{(\%)} & \textbf{(\%)} & \textbf{(min)} \\
\midrule
\multirow{2}{*}{Prithvi 100m} 
 & Full FT  & 8.39  & 4.78  & 5.85  & 7.36  & 3.61  & 3.47  & 5.78  & 72.82  & 4.51  & 0.41 \\
 & BitFit   & 6.88  & 4.68  & 5.68  & 7.40  & 3.55  & 3.13  & 4.24  & 72.11  & 4.51  & 0.43 \\
\midrule
\multirow{2}{*}{Prithvi 300m} 
 & Full FT  & 13.04 & 6.16  & 7.09  & 7.27  & 3.67  & 11.35 & 14.38 & 45.80  & 3.22  & 0.47 \\
 & BitFit   & 8.95  & 4.53  & 6.29  & 7.27  & 3.68  & 4.33  & 6.81  & 51.10  & 3.58  & 0.44 \\
\midrule
\multirow{2}{*}{Prithvi 600m} 
 & Full FT  & 38.94 & 14.21 & 14.08 & 6.90  & 3.67  & 14.72 & 18.51 & 52.95  & 2.68  & 0.88 \\
 & BitFit   & 18.47 & 9.84  & 14.24 & 6.74  & 3.63  & 10.69 & 11.14 & 57.90  & 2.86  & 0.87 \\
\midrule
\multirow{2}{*}{Croma}        
 & Full FT  & 36.94 & 10.27 & 22.80 & 7.23  & 4.02  & 4.32  & 5.76  & 71.41  & 3.02  & 1.40 \\
 & BitFit   & 31.42 & 5.51  & 20.25 & 7.13  & 4.45  & 4.32  & 5.61  & 76.09  & 3.30  & 1.34 \\ 
\midrule
\multirow{2}{*}{Dino-mm}      
 & Full FT  & 14.21 & 5.14  & 7.66  & 8.09  & 3.50  & 16.71 & 6.66  & 83.36  & 4.64  & 0.43 \\
 & BitFit   & 12.48 & 5.17  & 5.50  & 7.99  & 3.60  & 2.16  & 4.29  & 81.67  & 4.58  & 0.42 \\
\midrule
\multirow{2}{*}{Dofa}         
 & Full FT  & 12.58 & 5.15  & 8.13  & 7.19  & 3.64  & 3.27  & 3.91  & 63.64  & 3.12  & 0.59 \\
 & BitFit   & 9.00  & 4.89  & 8.21  & 7.23  & 3.72  & 2.72  & 3.45  & 70.62  & 3.31  & 0.54 \\
\midrule
\multirow{2}{*}{CMID}         
 & Full FT  & 21.68 & 4.07  & 12.68 & 7.14  & 3.92  & 9.90  & 3.81  & 78.77  & 3.68  & 0.78 \\
 & BitFit   & 21.57 & 4.02  & 12.04 & 7.23  & 3.79  & 8.14  & 3.22  & 77.22  & 3.44  & 0.74 \\
\midrule
\multirow{2}{*}{SARATR-X}     
 & Full FT  & 8.31  & 3.66  & 3.25  & 7.60  & 3.49  & 1.00  & 2.50  & 63.33  & 6.26  & 0.28 \\
 & BitFit   & 5.41  & 3.88  & 3.29  & 7.73  & 3.42  & 0.79  & 3.05  & 54.15  & 6.15  & 0.29 \\
\midrule
\multirow{2}{*}{FG-MAE}        
 & Full FT  & 11.45 & 3.99  & 4.76  & 7.05  & 3.70  & 6.76  & 14.63 & 38.87  & 3.98  & 0.48 \\
 & BitFit   & 7.31  & 3.84  & 4.28  & 7.27  & 3.63  & 4.28  & 12.65 & 34.45  & 3.74  & 0.51 \\
\midrule
\multirow{2}{*}{RVSA}         
 & Full FT  & 9.62  & 4.30  & 6.78  & 7.41  & 3.67  & 2.55  & 3.02  & 71.09  & 3.79  & 0.53 \\
 & BitFit   & 7.98  & 4.29  & 7.20  & 7.53  & 3.68  & 2.50  & 2.91  & 71.78  & 4.40  & 0.54 \\
\midrule
\multirow{2}{*}{SSL4EO}       
 & Full FT  & 4.21  & 3.64  & 2.19  & 7.73  & 3.31  & 1.56  & 4.76  & 27.69  & 5.94  & 0.21 \\
 & BitFit   & 3.44  & 3.96  & 1.98  & 7.15  & 3.31  & 1.39  & 4.48  & 51.54  & 7.22  & 0.21 \\
\bottomrule
\end{tabular}
}
\vspace{-5mm}
\end{table}

\subsubsection{Efficiency Metrics}
We report efficiency metrics for the full fine-tuning and BitFit strategies which provide an upper-bound  and lower bound reference for compute and memory (Table~\ref{tab:efficiency_metrics}); due to space limitations, results for other strategies are not included. BitFit demonstrates memory reductions across architectures. CROMA delivered the best overall F1 (0.761) under full fine-tuning, but at high training cost (36.94 GB) due to its 2D-ALiBi attention bias matrices and multimodal architecture overhead, though BitFit reduces this to 31.42 GB. Prithvi-600M performed best with LoRA (0.747), significantly reducing training memory needs while leveraging its large capacity, with BitFit showing even more dramatic savings (18.47 GB vs 38.94 GB). RVSA and Prithvi-300M also achieved their highest F1-scores with LoRA and VPT respectively, while BitFit provides substantial memory reductions for both models. Lightweight models SSL4EO and SARATR-X achieve their best performance (0.689, 0.713) under full fine-tuning while maintaining exceptional efficiency (4.21-8.31 GB training memory), with BitFit offering additional modest savings. In contrast, CMID and DINO-MM consistently underperformed across strategies, despite moderate parameter sizes. Overall, PEFT methods like LoRA and VPT often deliver near-optimal accuracy with substantially lower resource demands, suggesting a more scalable path for deploying FMs in remote sensing tasks.

\subsubsection{Transferability Across Time and Space}
\begin{figure}[htbp]
    \centering
    \includegraphics[width=\columnwidth]{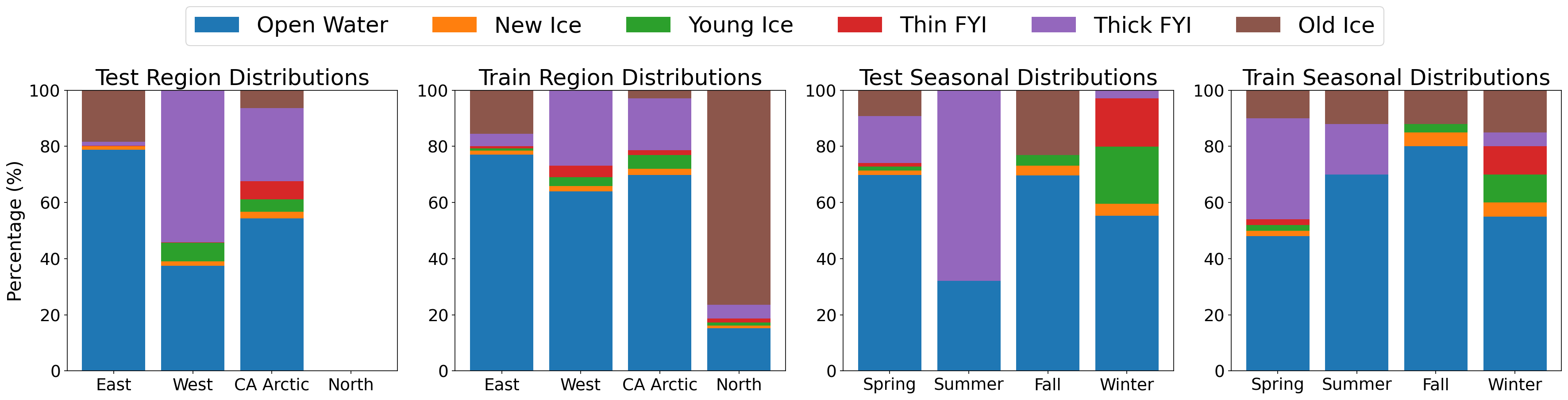}
    \caption{Distribution of ice types in the train and test data}
    \Description{Stacked bar charts showing the percentage distribution of ice types such as open water, new ice, young ice, thin first-year ice, thick first-year ice, and old ice in training and testing datasets across multiple regions and seasons. Some seasons contain higher proportions of young ice, while others have more old ice, indicating imbalance between categories.}
    \vspace{-4mm}
    \label{fig:test_distribution}
\end{figure}

To assess the transferability and robustness of FMs for sea ice type segmentation, we conducted comprehensive analyses examining seasonal transferability, spatial transferability, and performance scaling with training data size using the best-performing CROMA model.
\begin{figure}[htbp]
    \centering
    \includegraphics[width=\columnwidth]{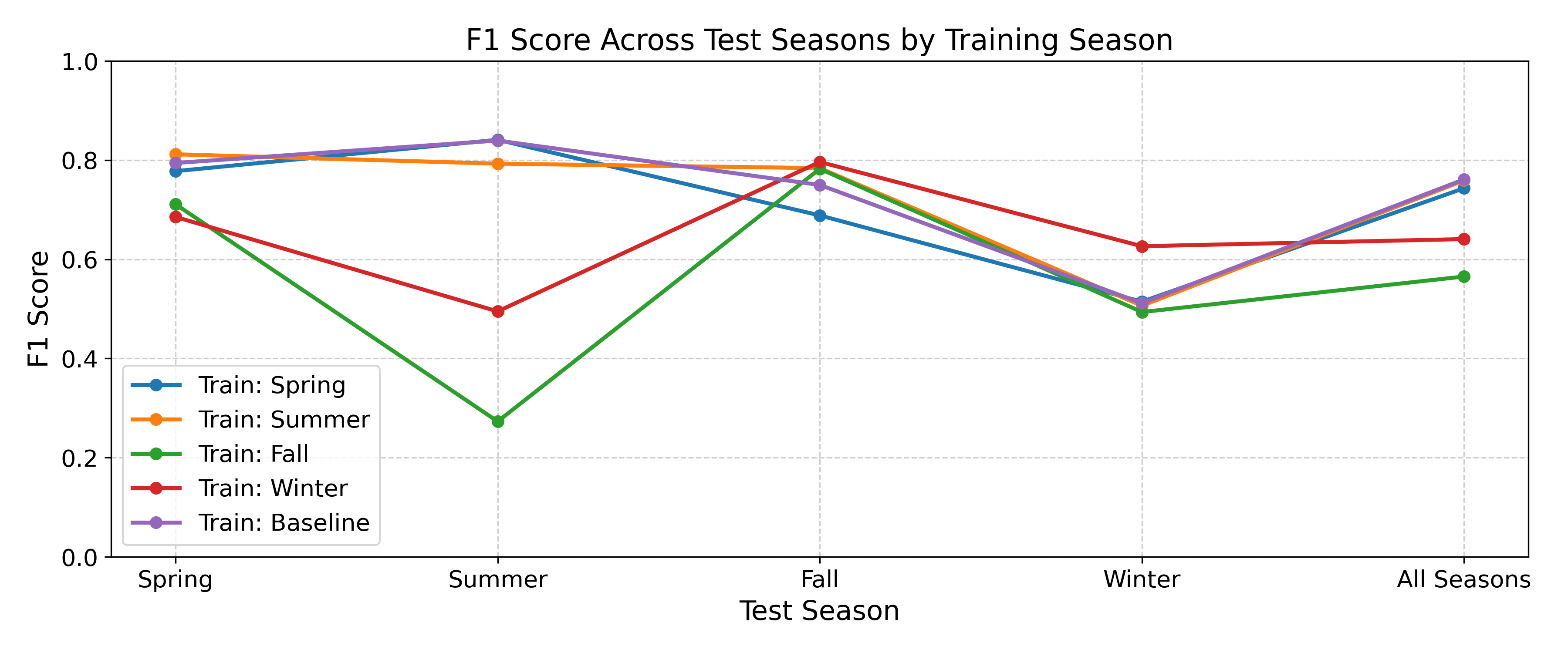}
    \caption{F1-score across test seasons for each training season}
    \Description{Line chart showing F1-scores when models trained in different seasons are evaluated across Spring, Summer, Fall, Winter, and All Seasons. Models trained in Fall perform the most consistently across all test seasons, while models trained in Summer drop significantly when tested on Fall. Training on All Seasons yields stable mid-range performance.}
    \vspace{-5mm}
    \label{fig:f1_across_seasons}
\end{figure}
Figure \ref{fig:f1_across_seasons} presents the F1-scores of the CROMA model trained on individual seasons and tested across all four seasons, as well as on the all-season dataset. The class distribution of ice types in the train and test set, shown in Figure \ref{fig:test_distribution}, provides context for performance variations. Generally, models performed best or achieved competitive scores when the test season matched the training season. When training on spring data, the model achieves strong performance on summer test conditions (0.84) and moderate performance on spring data itself (0.78), but experiences substantial degradation on fall and severe performance drops on winter conditions. Per-class analysis reveals that spring training provides superior summer performance for classes open water, thin FYI (Fist Year Ice), and especially class thick FYI. Summer-trained models demonstrate the most robust cross-seasonal transferability, maintaining relatively consistent F1-scores across spring, summer, and fall test seasons, with only winter showing significant degradation. Intriguingly, summer training also exhibits reciprocal cross-seasonal superiority, achieving better performance on spring data (0.81) than spring training achieves on its own seasonal data (0.78). Class-wise analysis shows summer training excels on spring test data for class open water while showing comparable or slightly better performance for classes thin and thick FYI. Fall training exhibits the most extreme seasonal bias, achieving reasonable performance on fall test data but catastrophic failure when applied to summer conditions, representing the worst cross-seasonal transfer observed. Performance on spring and winter test data also shows notable degradation, making fall-trained models highly specialized but poorly transferable. Winter training presents unexpected patterns, with moderate performance across spring, fall, and summer test seasons, but surprisingly modest performance on winter data itself. The presence of more young ice during winter increases the complexity of the prediction for the models.
\begin{figure}[htbp]
    \centering
    \includegraphics[width=\columnwidth]{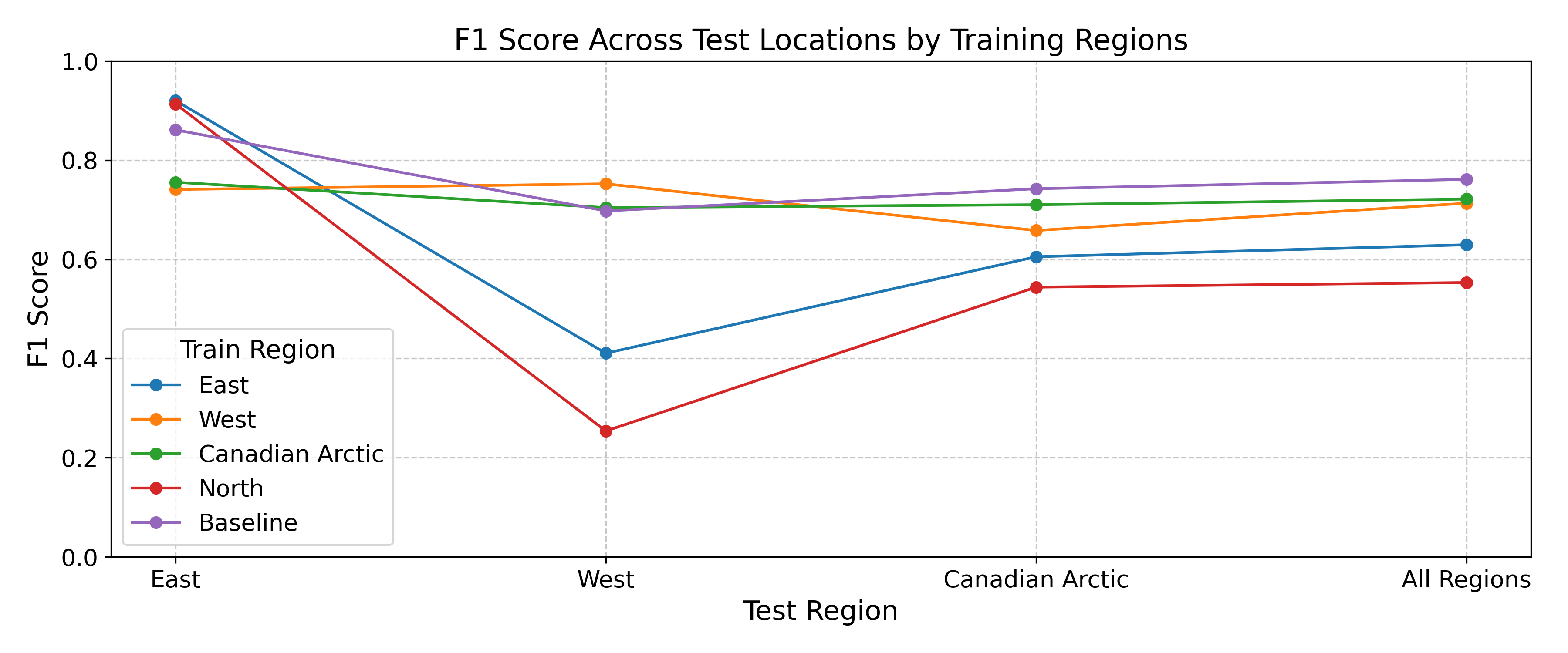}
    \caption{F1-scores by test location for each training location}
    \Description{Line chart showing F1-scores when models trained in one region, including East, West, and Canadian Arctic, are evaluated across multiple test regions. Models trained in East generalize best across regions, while models trained in West perform poorly when tested on Canadian Arctic. Training on all regions yields more stable results.}
    \vspace{-5mm}
    \label{fig:f1_spatial}
    \vspace{-3mm}
\end{figure}

\begin{figure}[htbp]
    \centering
    \includegraphics[ width=\linewidth]{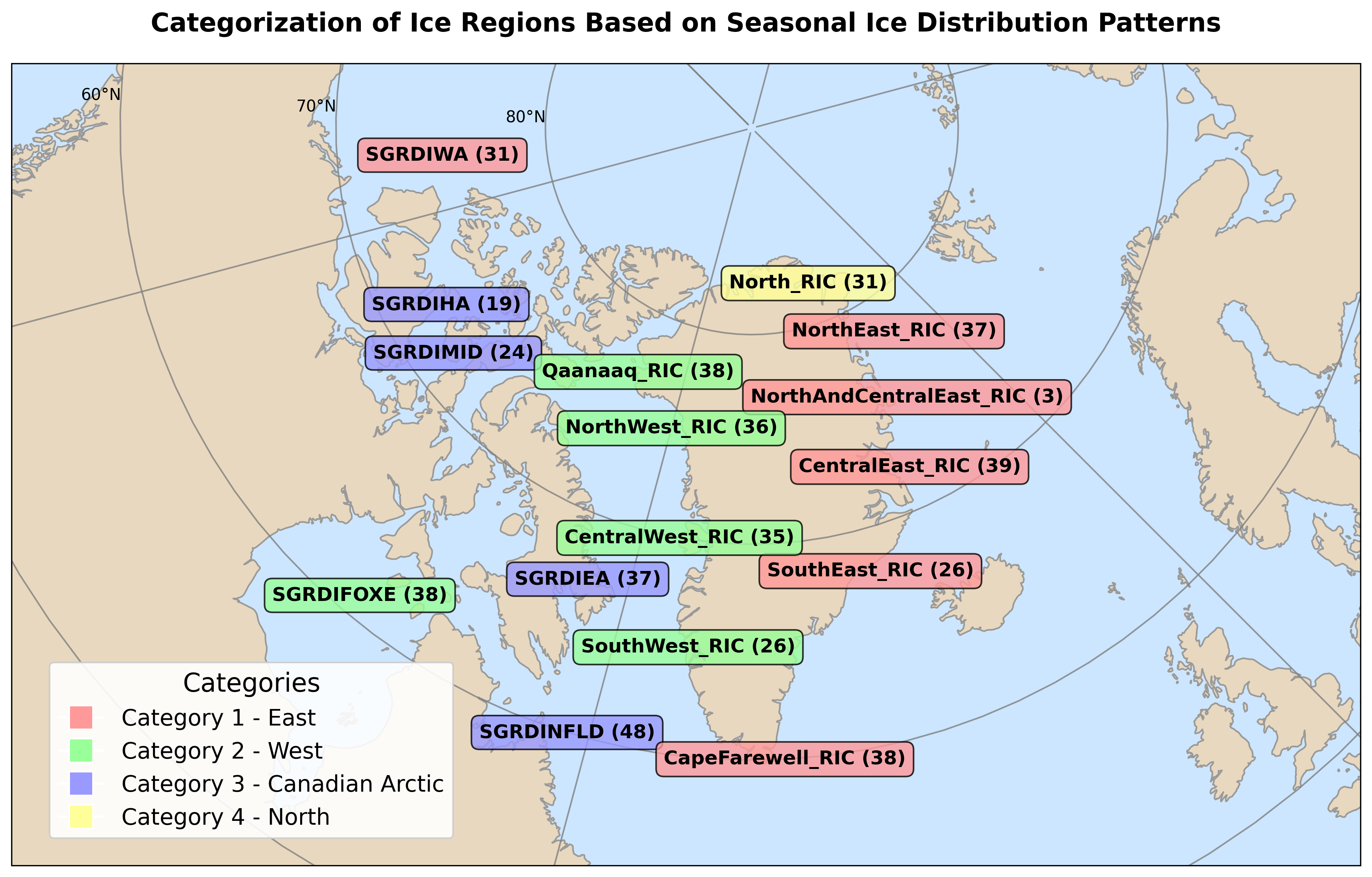} 
    \caption{Distribution of SAR training scenes categorized by ice charting region and season }
    \Description{Map of the Arctic showing SAR training scene locations grouped into regional categories labeled East, West, Central Arctic, and North. Each region is highlighted with colored boxes and tags indicating the number of scenes per location. Eastern regions such as Labrador and Greenland have the highest density, while Northern regions have fewer scenes.}

    \vspace{-5mm}
    \label{map}
\end{figure}

The spatial transferability of the CROMA model was examined by training it on data from one specific geographic region and testing its performance across other regions, as well as an aggregated "All Regions" dataset shown in Figure \ref{fig:f1_spatial}. The spatial transferability analysis demonstrates varying degrees of geographic transferability across different Arctic regions. To define these specific geographic regions, we analyzed 16 locations in the dataset monitored by CIS and DMI between 2018 and 2021, identifying four primary regional categories based on their characteristic seasonal ice class distributions. Figure \ref{map} illustrates this categorization and shows the distribution of training scenes within these defined ice charting regions and seasons. The numbers in parentheses denote the number of scenes within each region. 
Model trained on East data exhibit strong local performance, but show limited transferability to other regions, with performance dropping to 0.41 in the West and 0.61 in the Canadian Arctic. This suggests that ice characteristics in the East differ substantially from those in other regions. Models trained on West data demonstrate relatively stable performance across test regions, with F1-scores ranging from 0.66 to 0.75, indicating moderate spatial transferability. In contrast, Canadian Arctic-trained models do not show strong locality but transferable reasonably well to East and West regions. The all-region training approach (Baseline) achieves balanced performance across test regions, highlighting the benefit of incorporating geographic diversity to improve model robustness, though some regional discrepancies remain.

\vspace{-3mm}
\subsubsection{Data Size Sensitivity}
\begin{figure}[!t]
    \centering
    \includegraphics[width=\columnwidth]{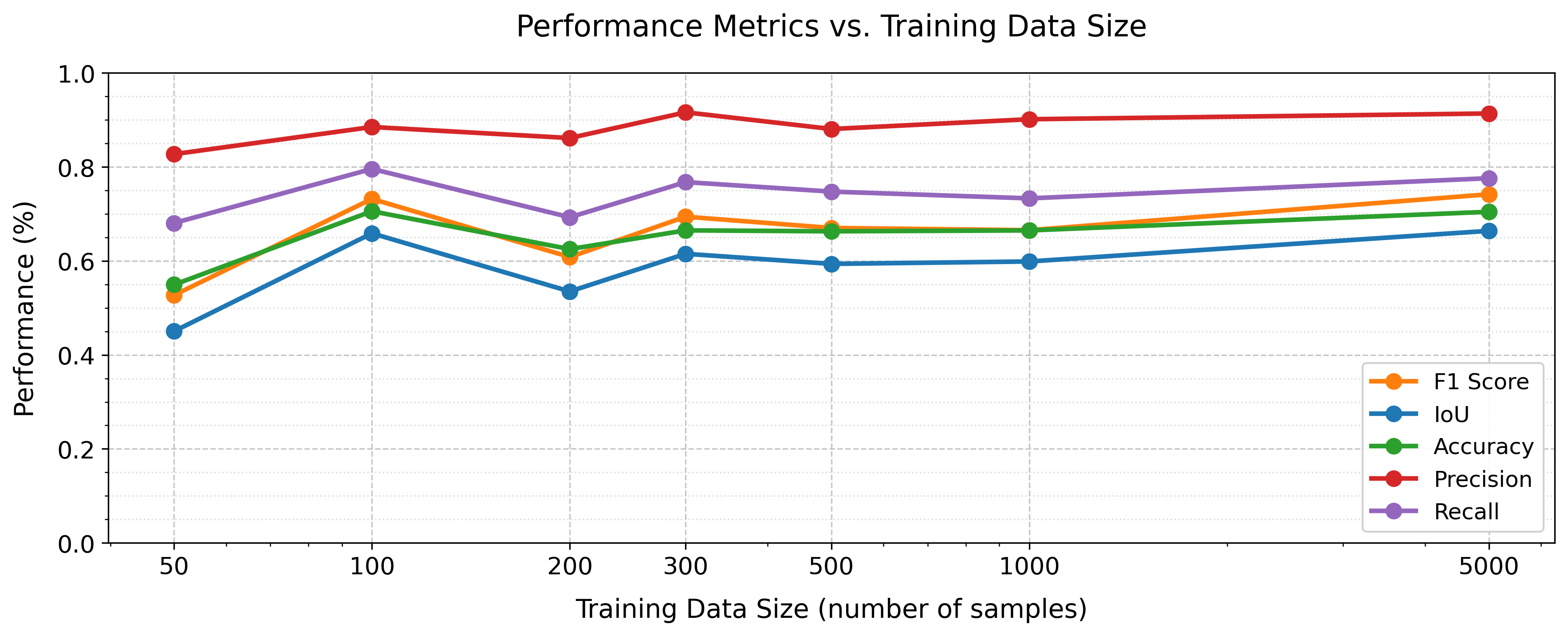}
    \caption{Performance metrics as a function of data size}
    \Description{Line chart showing performance metrics as training data size increases from 50 to 5000 samples. Metrics include F1 score, IoU, accuracy, precision, and recall. All metrics improve with more data, with recall consistently highest and IoU lowest. The steepest improvement occurs between 50 and 200 samples, after which gains plateau gradually.}

    \label{fig:datasize_metrics}
    \vspace{-3mm}
\end{figure}

Figure \ref{fig:datasize_metrics} illustrates how model performance varies with training dataset size. While a general trend of improved accuracy with more data is observed, the gains are not strictly linear. Notably, the 100-sample model performs surprisingly well, often matching or outperforming mid-range sizes (200–1000 samples), where performance fluctuates and even temporarily declines (e.g., F1 drops at 200). Performance appears to plateau or even decline in the mid-range (200-1000 samples), with metrics fluctuating rather than steadily improving. The largest dataset (5000 samples) achieves the highest F1-score and precision, but offers only modest improvement. These results suggest that around 500 samples may offer a practical balance between performance and data requirements, though additional gains are still achievable with larger datasets.

\begin{table}[t]
\centering
\caption{U-Net performance with and without KD}
\label{tab:unet_kd_comparison}
\setlength{\tabcolsep}{2pt}
\renewcommand{\arraystretch}{0.85}
\scriptsize
\begin{tabular}{lccccc}
\toprule
Model & F1 & Acc & Prec & Rec & IoU \\
\midrule
U-Net-Baseline & 0.766 & 0.743 & 0.899 & 0.743 & 0.696 \\
\textbf{U-Net-KD} & \textbf{0.787} & \textbf{0.771} & \textbf{0.851} & \textbf{0.771} & \textbf{0.727} \\
\bottomrule
\end{tabular}
\vspace{-5mm}
\end{table}

\section{Knowledge Distillation from Spatial and Seasonal Experts}
Despite the strong capabilities of FMs, our results show they do not consistently surpass baseline models in sea ice segmentation, indicating that their full potential for polar remote sensing applications is not fully leveraged. As demonstrated in our transferability experiments (Figures~\ref{fig:f1_across_seasons} and~\ref{fig:f1_spatial}), FMs tend to excel within the specific season or region they were trained on, acting as specialized ``experts``. This domain-specific proficiency suggests that while individual models learn specialized features, their expertise remains constrained to localized conditions rather than achieving broad generalization.

To overcome this limitation and consolidate the strengths of these varied expert models, we propose a multi-teacher knowledge distillation (KD) \cite{Hinton2015} method to train a single student model that learns from multiple spatiotemporally localized expert teacher models \cite{Zhang2023, Fukuda2017}. Specifically, our ensemble of teachers comprises of eight expert models, including four seasonal experts that are each trained on data from one of the four seasons, and four location experts each trained on data from one of the four distinct geographic regions. The knowledge distillation process involves training a unified student model that learns to replicate the combined expertise of specialized teachers. The student model receives supervision not only from ground truth labels but also from the soft predictions of each expert model. This approach allows the student model to capture the nuanced understanding that each expert developed for its specific environmental conditions, while learning to seamlessly integrate this knowledge for improved generalization.
Each expert teacher $T_i$ produces soft logits $z_i(\mathbf{x})$ on input patch $\mathbf{x}$. We train a student model $S$ to minimize a composite loss:
\[
\mathcal{L} \;=\; \alpha\,\mathcal{L}_{\mathrm{CE}}\bigl(S(\mathbf{x}), y\bigr)
\;+\;(1-\alpha)\,\frac{1}{M}
\sum_{i=1}^{M}
\mathrm{KL}\bigl(\sigma\bigl(z_i(\mathbf{x})/T\bigr)\,\Vert\,\sigma\bigl(S(\mathbf{x})/T\bigr)\bigr),
\]
where $\mathcal{L}_{\mathrm{CE}}$ is the standard cross-entropy to the hard labels $y$, KL denotes the
Kullback–Leibler divergence between teacher and student softmax distributions (temperature $T\!=\!3$), $M=8$ is the number of experts, and $\alpha=0.5$ balances the hard-label and distillation objectives. All teachers contribute equally to the soft distillation target.
We use a U-Net ([32, 32, 64, 64] channels) as the student and pretrained CROMA models as frozen teachers. The student is trained with AdamW ($1\times10^{-4}$ LR, $1\times10^{-2}$ weight decay) and CosineAnnealingLR, using a batch size of 32 and the same data augmentations as in the experimental pipeline.



Based on the knowledge distillation results presented in Table \ref{tab:unet_kd_comparison}, the consolidation of expert knowledge into a unified model demonstrates meaningful improvements across all evaluation metrics. The U-Net-KD student model achieves a higher F1-score (0.784 vs 0.76) compared to the baseline U-Net model and all fine-tuned FM models in Table \ref{tab:foundation_models_results}. These results suggest that incorporating soft supervision from multiple specialized FM expert models allows the student model to generalize better by learning a more nuanced and comprehensive representation of the underlying data distribution. Overall, knowledge distillation proves effective in transferring diverse domain-specific insights into a single robust segmentation model. While further adaptation is needed for polar applications, this demonstrates that existing FMs can be effectively leveraged for specialized tasks through knowledge distillation.
\section{Conclusions and Future Directions}
We introduced Ice-FMBench, a benchmark for evaluating foundation models (FMs) on SAR-based sea ice type segmentation. Using it, we compared multiple models across five fine-tuning strategies, jointly assessing accuracy and efficiency. Results show that FMs trained on lower-latitude remote sensing do not fully transfer to Arctic conditions due to differences in sensor modes and environmental dynamics. Larger models like Prithvi-600M outperform a U-Net baseline, while models such as CROMA become competitive under full fine-tuning.

Full fine-tuning remains effective for SAR-pretrained models, whereas LoRA offers the best balance of performance and efficiency for large FMs. VPT performs well on mid-sized models like Prithvi-300M, but minimal tuning yields limited benefit. To better adapt FMs to Arctic SAR, mitigating domain shift and overfitting is critical. Our knowledge distillation experiments demonstrate that aggregating multiple specialized models can improve generalization while remaining lightweight. Future work includes refining PEFT with mechanistic interpretability for more targeted adaptations.

\vspace{-3mm}
\begin{acks}
The authors would like to acknowledge the support of the U.S. National Science Foundation under Grants No. 2026962 and 2026865, and the University of Colorado Denver's Alderaan cluster for providing computational resources.
\end{acks}

\bibliographystyle{ACM-Reference-Format}
\bibliography{ref}


\end{document}